\documentclass[11pt,a4paper]{article}
\usepackage[hyperref]{acl2017}
\usepackage{times}
\usepackage{latexsym}

\usepackage{url}

\aclfinalcopy 
\def\aclpaperid{684} 

\usepackage{url}
\usepackage{times}
\usepackage{latexsym}
\usepackage{graphicx}
\usepackage{booktabs}
\usepackage{amsmath}
\usepackage{amssymb}
\usepackage{booktabs}
\usepackage{multirow}

\usepackage{caption}
\usepackage{subcaption}
\usepackage{soul}
\usepackage{float}
\usepackage{wrapfig}
\DeclareMathOperator{\bigru}{\overset{\longleftrightarrow}{\mathrm{GRU}}}
\usepackage{xcolor}
\definecolor{darkblue}{rgb}{0.0, 0.0, 0.55}
\hypersetup{citecolor=darkblue}

\title{Gated-Attention Readers for Text Comprehension}

\author{Bhuwan Dhingra\thanks{\enskip BD and HL contributed equally to this work.}\qquad Hanxiao Liu\footnotemark[1]\qquad Zhilin Yang\\ {\bf William W. Cohen \qquad Ruslan Salakhutdinov} \\
School of Computer Science\\
Carnegie Mellon University\\
\texttt{\{bdhingra,hanxiaol,zhiliny,wcohen,rsalakhu\}@cs.cmu.edu}}
\date{}

\begin{document}

\maketitle

\begin{abstract}
In this paper we study the problem of answering cloze-style questions over documents. Our model, the Gated-Attention (GA) Reader\footnote{Source code is available on github: \url{https://github.com/bdhingra/ga-reader}}, integrates a multi-hop architecture with a novel attention mechanism, which is based on multiplicative interactions between the query embedding and the intermediate states of a recurrent neural network document reader. This enables the reader to build query-specific representations of tokens in the document for accurate answer selection. The GA Reader obtains state-of-the-art results on three benchmarks for this task--the CNN \& Daily Mail news stories and the Who Did What dataset.
The effectiveness of multiplicative interaction is demonstrated by an ablation study,
and by comparing to alternative compositional operators for implementing the gated-attention. 
\end{abstract}

\section{Introduction}
A recent trend to measure progress towards machine reading is to test a system's ability to answer questions about a document it has to comprehend. Towards this end, several large-scale datasets of cloze-style questions over a context document have been introduced recently, which allow the training of supervised machine learning systems \citep{hermann2015teaching,hill2015goldilocks,onishi2016did}. Such datasets can be easily constructed automatically and the unambiguous nature of their queries provides an objective benchmark to measure a system's performance at text comprehension. 

Deep learning models have been shown to outperform traditional shallow approaches
on text comprehension tasks \citep{hermann2015teaching}.
The success of many recent models can be attributed primarily to two factors:
(1) \emph{Multi-hop architectures} 
\citep{weston2014memory, sordoni2016iterative, shen2016reasonet},
allow a model to scan the document and the question iteratively for multiple passes.
(2) \emph{Attention mechanisms},
\citep{chen2016thorough, hermann2015teaching}
borrowed from the machine translation literature \citep{bahdanau2014neural},
allow the model to focus on appropriate subparts of the context document.
Intuitively,
the multi-hop architecture allows the reader to incrementally refine token representations,
and the attention mechanism re-weights different parts in the document according to their relevance to the query.

The effectiveness of multi-hop reasoning and attentions have been explored orthogonally so far
in the literature.
In this paper,
we focus on combining both in a complementary manner,
by designing a novel attention mechanism which gates the evolving token representations across hops.
More specifically,
unlike existing models where the query attention
is applied either token-wise \citep{hermann2015teaching, kadlec2016text, chen2016thorough, hill2015goldilocks} or
sentence-wise \citep{weston2014memory, sukhbaatar2015end} to allow weighted aggregation,
the Gated-Attention (GA) module proposed in this work allows the query to directly interact with each dimension
of the token embeddings at the semantic-level,
and is applied layer-wise as information filters during the multi-hop representation learning process.
Such a fine-grained attention enables our model to learn conditional token representations w.r.t.\ the given question,
leading to accurate answer selections. 

We show in our experiments that the proposed GA reader,
despite its relative simplicity,
consistently improves over a variety of strong baselines on three benchmark datasets
. 
Our key contribution, the GA module, provides a significant improvement for large datasets. Qualitatively, visualization of the attentions at intermediate layers of the GA reader shows that in each layer the GA reader attends to distinct salient aspects of the query which help in determining the answer. 

\section{Related Work}
\label{sec:related}
The cloze-style QA task involves tuples of the form $(d,q,a,\mathcal{C})$, where $d$ is a document (context), $q$ is a query over the contents of $d$, in which a phrase is replaced with a placeholder, and $a$ is the answer to $q$, which comes from a set of candidates $\mathcal{C}$. In this work we consider datasets where each candidate $c \in \mathcal{C}$ has at least one token which also appears in the document. The task can then be described as: given a document-query pair $(d,q)$, find $a \in \mathcal{C}$ which answers $q$. Below we provide an overview of representative neural network architectures which have been applied to this problem.

\textit{LSTMs with Attention:} Several architectures introduced in \citet{hermann2015teaching}
employ LSTM units to compute a combined document-query representation $g(d,q)$,
which is used to rank the candidate answers.
These include the \textbf{DeepLSTM Reader} which performs a single forward pass through the concatenated \textit{(document, query)} pair to obtain $g(d,q)$;
the \textbf{Attentive Reader} which first computes a document vector $d(q)$ by a weighted aggregation of words according to attentions based on $q$, and then combines $d(q)$ and $q$ to obtain their joint representation $g(d(q),q)$;
and the \textbf{Impatient Reader} where the document representation is built incrementally.
The architecture of the Attentive Reader has been simplified recently in \textbf{Stanford Attentive Reader},
where shallower recurrent units were used with a bilinear form for the query-document attention \citep{chen2016thorough}.

\textit{Attention Sum:} The \textbf{Attention-Sum (AS) Reader}
\citep{kadlec2016text} uses two bi-directional GRU networks \citep{cho2014learning} to encode both $d$ and $q$ into vectors.
A probability distribution over the entities in $d$
is obtained by computing dot products between $q$ and the entity embeddings and taking a softmax. 
Then, an aggregation scheme named \emph{pointer-sum attention}
is further applied to sum the probabilities of the same entity,
so that frequent entities the document will be favored
compared to rare ones. 
Building on the AS Reader,
the \textbf{Attention-over-Attention (AoA) Reader} \citep{cui2016attention} 
introduces a two-way attention mechanism where the query and the document are mutually attentive to each other.

\textit{Mulit-hop Architectures:} \textbf{Memory Networks (MemNets}) were proposed in \citet{weston2014memory},
where each sentence in the document is encoded to a memory by aggregating nearby words. 
Attention over the memory slots given the query is used to compute an overall memory and to renew the query representation over multiple iterations, allowing certain types of reasoning over the salient facts in the memory and the query. 
\textbf{Neural Semantic Encoders (NSE)} \citep{munkhdalai2016neural} extended MemNets by introducing a \textit{write} operation which can evolve the memory over time during the course of reading.
Iterative reasoning has been found effective in several more recent models,
including the \textbf{Iterative Attentive Reader} \citep{sordoni2016iterative} and \textbf{ReasoNet} \citep{shen2016reasonet}.
The latter allows dynamic reasoning steps
and is trained with reinforcement learning. 


Other related works include \textbf{Dynamic Entity Representation network (DER}) \citep{kobayashi2016dynamic},
which builds dynamic representations of the candidate answers while reading the document, and accumulates the information about an entity by max-pooling;
\textbf{EpiReader} \citep{trischler2016natural}
consists of two networks, where one proposes a small set of candidate answers, and the other reranks the proposed candidates conditioned on the query and the context; \textbf{Bi-Directional Attention Flow network (BiDAF)} \cite{seo2016bidirectional} adopts a multi-stage hierarchical architecture along with a flow-based attention mechanism;
\citet{bajgar2016embracing} showed a 10\% improvement on the CBT corpus \citep{hill2015goldilocks} by training the AS Reader on an augmented training set of about 14 million examples, making a case for the community to exploit data abundance. The focus of this paper, however, is on designing models which exploit the available data efficiently.

\section{Gated-Attention Reader}
\label{sec:ga}
Our proposed GA readers perform multiple hops over the document (context), similar to the Memory Networks architecture \citep{sukhbaatar2015end}.
Multi-hop architectures mimic the multi-step comprehension process of human readers,
and have shown promising results in several recent models for text comprehension \citep{sordoni2016iterative, kumar2015ask, shen2016reasonet}.
The contextual representations in GA readers,
namely the embeddings of words in the document,
are iteratively refined across hops
until reaching a final attention-sum module \citep{kadlec2016text} which maps 
the contextual representations in the last hop to a probability distribution over candidate answers.

The attention mechanism has been introduced recently to model human focus,
leading to significant improvement in machine translation and image captioning \citep{bahdanau2014neural, mnih2014recurrent}.
In reading comprehension tasks, ideally,
the semantic meanings carried by the contextual embeddings should be aware of the query across hops.
As an example,
human readers are able to keep the question in mind during multiple passes of reading,
to successively mask away information irrelevant to the query.
However,
existing neural network readers are restricted to either attend to tokens \citep{hermann2015teaching, chen2016thorough}
or entire sentences \citep{weston2014memory}, with the assumption that certain sub-parts of the document are more important than others.
In contrast, 
we propose a finer-grained model which attends to components of the semantic representation being built up by the GRU.
The new attention mechanism,
called \emph{gated-attention},
is implemented via \emph{multiplicative} interactions between the query and the contextual embeddings,
and is applied per hop to act as fine-grained information filters during the multi-step reasoning. The filters weigh individual components of the vector representation of \textit{each} token in the document separately.

The design of gated-attention layers is motivated by the effectiveness of multiplicative interaction
among vector-space representations,
e.g., in various types of recurrent units \citep{hochreiter1997long, wu2016multiplicative} and in relational learning \citep{yang2014learning, kiros2014multiplicative}.
While other types of compositional operators are possible,
such as concatenation or addition \citep{mitchell2008vector},
we find that multiplication has strong empirical performance
(section \ref{sec:ablation}),
where query representations naturally serve as information filters
across hops.

\subsection{Model Details}
\label{sec:model-details}
Several components of the model use a Gated Recurrent Unit (GRU) \citep{cho2014learning} which maps an input sequence $X = [x_1,x_2,\ldots,x_T]$ to an ouput sequence $H=[h_1,h_2,\ldots,h_T]$ as follows:
\begin{align*}
r_t &= \sigma (W_r x_t + U_r h_{t-1} + b_r), \\
z_t &= \sigma (W_z x_t + U_z h_{t-1} + b_z), \\
\tilde{h}_t &= \tanh(W_h x_t + U_h (r_t \odot h_{t-1}) + b_h), \\
h_t &= (1-z_t) \odot h_{t-1} + z_t \odot \tilde{h}_t.
\end{align*}
where $\odot$ denotes the Hadamard product or the element-wise multiplication.
$r_t$ and $z_t$ are called the \textit{reset} and \textit{update} gates respectively, and $\tilde{h}_t$ the \textit{candidate output}. 
A Bi-directional GRU (Bi-GRU) processes the sequence in both forward and backward directions to produce two sequences $[h_1^f, h_2^f, \dots, h_T^f ]$ and $[ h_1^b, h_2^b, \dots, h_T^b ]$, which are concatenated at the output
\begin{equation}
    \bigru(X) = [h^f_1 \Vert h^b_T,\ldots,h^f_T \Vert h^b_1]
\end{equation}
where $\bigru(X)$ denotes the \textit{full output} of the Bi-GRU obtained by concatenating each forward state $h^f_i$ and backward state $h^b_{T-i+1}$ at step $i$ given the input $X$. Note $\bigru(X)$ is a matrix in $\mathbb{R}^{2n_h \times T}$ where $n_h$ is the number of hidden units in GRU.

Let $X^{(0)} = [x_1^{(0)},x_2^{(0)},\ldots x_{|D|}^{(0)}]$ denote the token embeddings of the document, which are also inputs at layer 1 for the document reader below, and $Y = [y_1,y_2,\ldots y_{|Q|}]$ denote the token embeddings of the query. Here $|D|$ and $|Q|$ denote the document and query lengths respectively.


\begin{figure*}[t]
\centering
\caption{Gated-Attention Reader. Dashed lines represent dropout connections.}
\includegraphics[width=\linewidth,trim={0 25mm 0 25mm},clip]{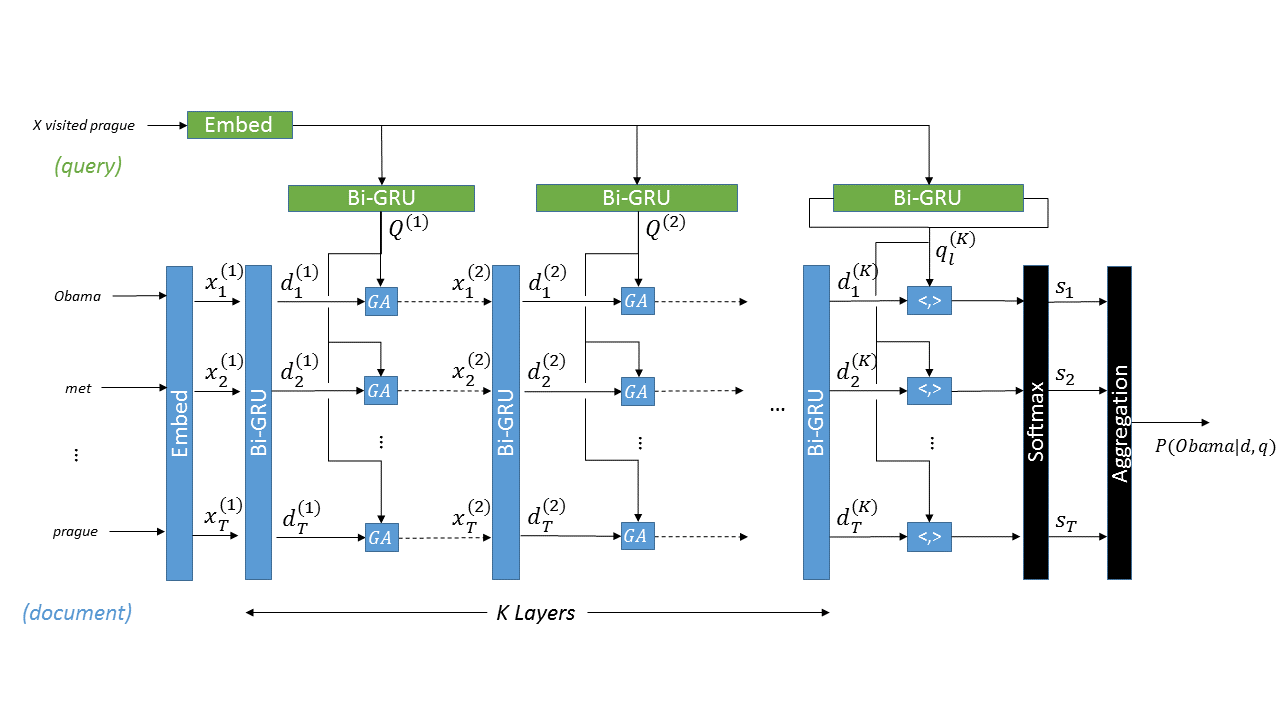}
\label{fig:model}
\end{figure*}

\subsubsection{Multi-Hop Architecture}
Fig.\ \ref{fig:model} illustrates the Gated-Attention (GA) reader. 
The model reads the document and the query over $K$ horizontal layers,
where layer $k$ receives the contextual embeddings $X^{(k-1)}$ of the document from the previous layer. The document embeddings are transformed by taking the full output of a document Bi-GRU (indicated in blue in Fig.\ \ref{fig:model}):
\begin{equation}
    D^{(k)} = \bigru^{(k)}_D(X^{(k-1)})
\end{equation}
At the same time, a layer-specific query representation is computed as the full output of a separate query Bi-GRU (indicated in green in Figure \ref{fig:model}):
\begin{equation}
    Q^{(k)} = \bigru^{(k)}_Q(Y)
\end{equation}

Next, \textit{Gated-Attention} is applied to $D^{(k)}$ and $Q^{(k)}$ to compute inputs for the next layer $X^{(k)}$.
\begin{equation}
    X^{(k)} = \mathrm{GA}(D^{(k)}, Q^{(k)})
\end{equation}
where GA is defined in the following subsection.

\subsubsection{Gated-Attention Module}
For brevity,
let us drop the superscript $k$ in this subsection as we are focusing on a particular layer.
For each token $d_i$ in $D$,
the GA module forms a token-specific representation of the query $\tilde{q}_i$ using soft attention, and then multiplies the query representation element-wise with the document token representation. Specifically, for $i=1,\ldots,|D|$:
\begin{align}
    \alpha_i &= \text{softmax}(Q^\top d_i) \label{eq:q_att} \\
\tilde{q}_i &= Q\alpha_i \nonumber \\
x_i &= d_i \odot \tilde{q}_i
\label{eq:gating}
\end{align}
In equation \eqref{eq:gating} we use the multiplication operator
to model the interactions between $d_i$ and $\tilde{q}_i$.
In the experiments section,
we also report results for other choices of gating functions,
including addition $x_i = d_i + \tilde{q}_i$ and concatenation $x_i = d_i \Vert \tilde{q}_i$.



\subsubsection{Answer Prediction}
Let $q^{(K)}_\ell = q_\ell^f\Vert q^b_{T-\ell+1}$ be an intermediate output of the final layer query Bi-GRU at the location $\ell$ of the cloze token in the query,
and $D^{(K)} = \bigru^{(K)}_D(X^{(K-1)})$ be the full output of final layer document Bi-GRU. To obtain the probability that a particular token in the document answers the query, we take an inner-product between these two, and pass through a softmax layer:
\begin{equation}
\label{eq:att}
s = \text{softmax}((q^{(K)}_\ell)^T D^{(K)})
\end{equation}
where vector $s$ defines a probability distribution over the $|D|$ tokens in the document. The probability of a particular candidate $c \in \mathcal{C}$ as being the answer is then computed by aggregating the probabilities of all document tokens which appear in $c$ and renormalizing over the candidates:
\begin{equation}
    \Pr(c|d,q) \propto \sum_{i\in \mathbb{I}(c,d)} s_i
\end{equation}
where $\mathbb{I}(c,d)$ is the set of positions where a token in $c$ appears in the document $d$. This aggregation operation is the same as the \textit{pointer sum attention} applied in the AS Reader \citep{kadlec2016text}.

Finally, the candidate with maximum probability is selected as the predicted answer:
\begin{equation}
    a^* = \mathrm{argmax}_{c \in \mathcal{C}} \enskip \Pr(c|d,q).
\end{equation}

During the training phase,
model parameters of GA are updated w.r.t.\ a cross-entropy loss between the predicted probabilities and the true answers.

\subsubsection{Further Enhancements}
\label{sec:tricks}
\emph{Character-level Embeddings}: Given a token $w$ from the document or query, its vector space representation is computed as $x=L(w) || C(w)$. $L(w)$ retrieves the word-embedding for $w$ from a lookup table $L \in \mathbb{R}^{|V|\times n_l}$, whose rows hold a vector for each unique token in the vocabulary. We also utilize a character composition model $C(w)$ which generates an orthographic embedding of the token. Such embeddings have been previously shown to be helpful for tasks like Named Entity Recognition \citep{yang2016multi} and dealing with OOV tokens at test time \citep{dhingra2016tweet2vec}. The embedding $C(w)$ is generated by taking the final outputs $z^f_{n_c}$ and $z^b_{n_c}$ of a Bi-GRU applied to embeddings from a lookup table of characters in the token, and applying a linear transformation:
\begin{align*}
z &= z^f_{n_c} || z^b_{n_c}\\
C(w) &= W z + b
\end{align*}

\emph{Question Evidence Common Word Feature (qe-comm)}: \citet{li2016dataset} recently proposed a simple token level indicator feature which significantly boosts reading comprehension performance in some cases. For each token in the document we construct a one-hot vector $f_i \in \{0,1\}^2$ indicating its presence in the query. It can be incorporated into the GA reader by assigning a feature lookup table $F \in \mathbb{R}^{n_F \times 2}$ (we use $n_F=2$), taking the feature embedding $e_i=f_i^T F$ and appending it to the inputs of the last layer document BiGRU as, $x_i^{(K)} \Vert f_i$ for all $i$. We conducted several experiments both with and without this feature and observed some interesting trends, which are discussed below. Henceforth, we refer to this feature as the \textit{qe-comm feature} or just \textit{feature}.

\section{Experiments and Results}
\label{sec:results}
\subsection{Datasets}

We evaluate the GA reader on five large-scale datasets recently proposed in the literature. The first two, CNN and Daily Mail news stories\footnote{\scriptsize \url{https://github.com/deepmind/rc-data}} consist of articles from the popular CNN and Daily Mail websites \citep{hermann2015teaching}. A query over each article is formed by removing an entity from the short summary which follows the article. Further, entities within each article were anonymized to make the task purely a comprehension one. N-gram statistics, for instance, computed over the entire corpus are no longer useful in such an anonymized corpus.

The next two datasets are formed from two different subsets of the Children's Book Test (CBT)\footnote{\scriptsize \url{http://www.thespermwhale.com/jaseweston/babi/CBTest.tgz}} \citep{hill2015goldilocks}. Documents consist of 20 contiguous sentences from the body of a popular children's book, and queries are formed by deleting a token from the 21\textsuperscript{st} sentence. We only focus on subsets where the deleted token is either a common noun (CN) or named entity (NE) since simple language models already give human-level performance on the other types (cf. \citep{hill2015goldilocks}).

The final dataset is Who Did What\footnote{\scriptsize \url{ https://tticnlp.github.io/who_did_what/}} (WDW) \citep{onishi2016did},
constructed from the LDC English Gigaword newswire corpus.
First, article pairs which appeared around the same time and with overlapping entities are chosen, and then one article forms the document and a cloze query is constructed from the other. Missing tokens are always person named entities.
Questions which are easily answered by simple baselines are filtered out,
to make the task more challenging. There are two versions of the training set---a small but focused ``Strict'' version and a large but noisy ``Relaxed'' version.
We report results on both settings which share the same validation and test sets. Statistics of all the datasets used in our experiments are summarized in the Appendix (Table~\ref{tab:data}).

\subsection{Performance Comparison}

\begin{table*}[ht]
\parbox{.60\linewidth}{
\centering
\caption{\small Validation/Test accuracy (\%) on WDW dataset for both ``Strict'' and ``Relaxed'' settings. Results with ``$\dagger$'' are cf previously published works.}
\label{tab:wdw}
\begin{tabular}{l|cc|cc}
\toprule
\multirow{2}{*}{\textbf{Model}}  & \multicolumn{2}{c|}{\textbf{Strict}}              & \multicolumn{2}{c}{\textbf{Relaxed}}              \\ \cmidrule(l){2-5} 
                        & \multicolumn{1}{c|}{Val} & Test          & \multicolumn{1}{c|}{Val} & Test          \\ \midrule
Human  $\dagger$                 & --                       & 84          & --                       & --            \\ \midrule
Attentive Reader   $\dagger$     & --                       & 53          & --                       & 55         \\
AS Reader   $\dagger$            & --                       & 57          & --                       & 59          \\
Stanford AR    $\dagger$         & --                       & 64          & --                       & 65          \\
NSE $\dagger$& 66.5                     & 66.2          & 67.0                     & 66.7          \\ \midrule
GA-{}-  $\dagger$             & --                     & 57          & --                     & 60.0          \\
GA (update $L(w)$)              & 67.8                     & 67.0          & 67.0                     & 66.6          \\
GA (fix $L(w)$)              & 68.3                     & 68.0          & 69.6                     & 69.1          \\
GA (+feature, update $L(w)$)    & 70.1            & 69.5 & 70.9            & 71.0 \\
GA (+feature, fix $L(w)$)    & \textbf{71.6}            & \textbf{71.2} & \textbf{72.6}            & \textbf{72.6} \\\bottomrule
\end{tabular}
}
\hfill
\parbox{.375\linewidth}{
\centering
\caption{\small \textbf{Top:} Performance of different gating functions. \textbf{Bottom:} Effect of varying the number of hops $K$. Results on WDW without using the qe-comm feature and with fixed $L(w)$.}
\label{tab:gating_fn}
\begin{tabular}{@{}l|c|c@{}}
\toprule
\multirow{2}{*}{\textbf{Gating Function}} & \multicolumn{2}{c}{\textbf{Accuracy}} \\ \cmidrule(l){2-3} 
                                          & Val                & Test              \\ \midrule
Sum                                       & 64.9               & 64.5             \\
Concatenate                               & 64.4               & 63.7              \\
Multiply                                  & \textbf{68.3}               & \textbf{68.0}              \\ \midrule

$\mathbf{K}$	&	&	\\ \midrule
1 (AS) $\dagger$                   & --               & 57             \\
2			                               & 65.6               & 65.6              \\
3                                & \textbf{68.3}               & 68.0              \\
4                                & \textbf{68.3}               & \textbf{68.2}              \\\bottomrule
\end{tabular}
}
\end{table*}

\begin{table*}[ht]
\caption{\small Validation/Test accuracy $(\%)$ on CNN, Daily Mail and CBT.
Results marked with ``$\dagger$'' are cf previously published works.
Results marked with ``$\ddagger$'' were obtained by training on a larger training set. Best performance on standard training sets is in bold, and on larger training sets in italics.}
\label{tab:results}
\centering
\begin{tabular}{@{}l|cc|cc|cc|cc@{}}
\toprule
\multirow{2}{*}{\textbf{Model}}       & \multicolumn{2}{c|}{\textbf{CNN}}                     & \multicolumn{2}{c|}{\textbf{Daily Mail}}              & \multicolumn{2}{c|}{\textbf{CBT-NE}}           & \multicolumn{2}{c}{\textbf{CBT-CN}}          \\ \cmidrule(l){2-9} 
                                      & Val                      & Test                      & Val                      & Test                      & Val                   & Test                   & Val                   & Test                  \\ \midrule
Humans (query) $\dagger$                        & --                        & --                         & --                        & --                         & --                     & 52.0                   & --                     & 64.4                  \\
Humans (context + query) $\dagger$              & --                        & --                         & --                        & --                         & --                     & 81.6                   & --                     & 81.6                  \\ \midrule
LSTMs (context + query) $\dagger$               & --                        & --                         & --                        & --                         & 51.2                  & 41.8                   & 62.6                  & 56.0                  \\ 
Deep LSTM Reader $\dagger$ & 55.0 & 57.0 & 63.3 & 62.2 & -- & -- & -- & -- \\
Attentive Reader $\dagger$                      & 61.6                     & 63.0                      & 70.5                     & 69.0                      & --                     & --                      & --                     & --                     \\
Impatient Reader $\dagger$                      & 61.8                     & 63.8                      & 69.0                     & 68.0                      & --                     & --                      & --                     & --                     \\ 
MemNets $\dagger$                  & 63.4                     & 66.8                      & --                        & --                         & 70.4                  & 66.6                   & 64.2                  & 63.0                  \\
AS Reader $\dagger$              & 68.6                     & 69.5                      & 75.0                     & 73.9                      & 73.8                  & 68.6                   & 68.8                  & 63.4                  \\
DER Network $\dagger$                                   & 71.3                     & 72.9                      & --                        & --                         & --                     & --                      & --                     & --                     \\ 
Stanford AR (relabeling) $\dagger$	& 	73.8	&	73.6	&	77.6	&	76.6	&	--	&	--	&	--	&	--	\\
Iterative Attentive Reader $\dagger$	&	72.6	&	73.3	&	--	&	--	&	75.2	&	68.6	&	72.1	&	69.2 \\
EpiReader $\dagger$	&	73.4	&	74.0	&	--	&	--	&	75.3	&	69.7	&	71.5	&	67.4 \\
AoA Reader $\dagger$ &	73.1	&	74.4	&	--	&	--	&	77.8	&	72.0	&	72.2	&	69.4	\\ 
ReasoNet $\dagger$ &	72.9	&	74.7	&	77.6	&	76.6	&	--	&	--	&	--	&	--	\\
NSE $\dagger$ &	--	&	--	&	--	&	--	&	78.2	&	73.2	&	74.3	&	\textbf{71.9} \\
BiDAF $\dagger$ &	76.3 & 76.9 & 80.3 & 79.6 &	--	&	--	&	--	&	--	\\ \midrule
MemNets (ensemble) $\dagger$                    & 66.2                     & 69.4                      & --                        & --                         & --                     & --                      & --                     & --                     \\ 
AS Reader (ensemble) $\dagger$                  & 73.9                     & 75.4                      & 78.7                     & 77.7             & 76.2                  & 71.0                   & 71.1                  & 68.9                  \\ 
Stanford AR (relabeling,ensemble) $\dagger$	& 	77.2	&	77.6	&	80.2	&	79.2	&	--	&	--	&	--	&	--	\\ 
Iterative Attentive Reader (ensemble) $\dagger$	&	75.2	&	76.1	&	--	&	--	&	76.9	&	72.0	&	74.1	&	71.0 \\ 
EpiReader (ensemble) $\dagger$	&	--	&	--	&	--	&	--	&	76.6	&	71.8	&	73.6	&	70.6 \\ \midrule
AS Reader (+BookTest) $\dagger$ $\ddagger$ &	--	&	--	&	--	&	--	&	80.5	&	76.2	&	83.2	&	80.8	\\
AS Reader (+BookTest,ensemble) $\dagger$ $\ddagger$ &	--	&	--	&	--	&	--	&	\textit{82.3}	&	\textit{78.4}	&	\textit{85.7}	&	\textit{83.7}	\\ \midrule
GA-{}-   & 73.0   & 73.8    & 76.7     & 75.7    & 74.9     & 69.0  & 69.0      & 63.9         \\
GA (update $L(w)$)	&	\textbf{77.9}	&	\textbf{77.9}	&	\textbf{81.5}	&	\textbf{80.9}	&	76.7	&	70.1	&	69.8	&	67.3	\\
GA (fix $L(w)$)	&	77.9	&	77.8	&	80.4	&	79.6	&	77.2	&	71.4	&	71.6	&	68.0	\\
GA (+feature, update $L(w)$)	&	77.3	&	76.9	&	80.7	&	80.0	&	77.2	&	73.3	&	73.0	&	69.8	\\ 
GA (+feature, fix $L(w)$)	&	76.7	&	77.4	&	80.0	&	79.3	&	\textbf{78.5}	&	\textbf{74.9}	&	\textbf{74.4}	&	70.7	\\ \bottomrule
\end{tabular}
\end{table*}

Tables~\ref{tab:wdw} and \ref{tab:results} show a comparison of the performance of GA Reader with previously published results on WDW and CNN, Daily Mail, CBT datasets respectively. The numbers reported for GA Reader are for single best models, though we compare to both ensembles and single models from prior work. GA Reader-{}- refers to an earlier version of the model, unpublished but described in a preprint,
with the following differences---(1) it does not utilize token-specific attentions within the GA module, as described in equation \eqref{eq:q_att}, (2) it does not use a character composition model, (3) it is initialized with word embeddings pretrained on the corpus itself rather than GloVe. A detailed analysis of these differences is studied in the next section. Here we present 4 variants of the latest GA Reader, using combinations of whether the qe-comm feature is used (+feature) or not, and whether the word lookup table $L(w)$ is updated during training or fixed to its initial value. Other hyperparameters are listed in Appendix~\ref{app:implementation}.

Interestingly, we observe that feature engineering leads to significant improvements for WDW and CBT datasets, but not for CNN and Daily Mail datasets. We note that anonymization of the latter datasets means that there is already some feature engineering (it adds hints about whether a token is an entity), and these are much larger than the other four. In machine learning it is common to see the effect of feature engineering diminish with increasing data size. Similarly, fixing the word embeddings provides an improvement for the WDW and CBT, but not for CNN and Daily Mail. This is not surprising given that the latter datasets are larger and less prone to overfitting.

Comparing with prior work, on the WDW dataset the basic version of the GA Reader outperforms all previously published models when trained on the Strict setting. By adding the qe-comm feature the performance increases by 3.2\% and 3.5\% on the Strict and Relaxed settings respectively to set a new state of the art on this dataset. On the CNN and Daily Mail datasets the GA Reader leads to an improvement of 3.2\% and 4.3\% respectively over the best previous single models. They also outperform previous ensemble models, setting a new state of that art for both datasets. For CBT-NE, GA Reader with the qe-comm feature outperforms all previous single and ensemble models except the AS Reader trained on the much larger BookTest Corpus \citep{bajgar2016embracing}. Lastly, on CBT-CN the GA Reader with the qe-comm feature outperforms all previously published single models except the NSE, and AS Reader trained on a larger corpus. For each of the 4 datasets on which GA achieves the top performance, we conducted one-sample proportion tests to test whether GA is significantly better than the second-best baseline. The p-values are $0.319$ for CNN, $<$$0.00001$ for DailyMail, $0.028$ for CBT-NE, and $<$$0.00001$ for WDW. In other words, GA statistically significantly outperforms all other baselines on 3 out of those 4 datasets at the $5\%$ significance level. The results could be even more significant under paired tests, however we did not have access to the predictions from the baselines.

\subsection{GA Reader Analysis}
\label{sec:ablation}
\begin{figure*}[ht]
    \centering
    \caption{Performance in accuracy with and without the Gated-Attention module over different training sizes. $p$-values for an exact one-sided Mcnemar's test are given inside the parentheses for each setting.}
    \begin{subfigure}[b]{0.245\textwidth}
    \includegraphics[width=1.0\linewidth]{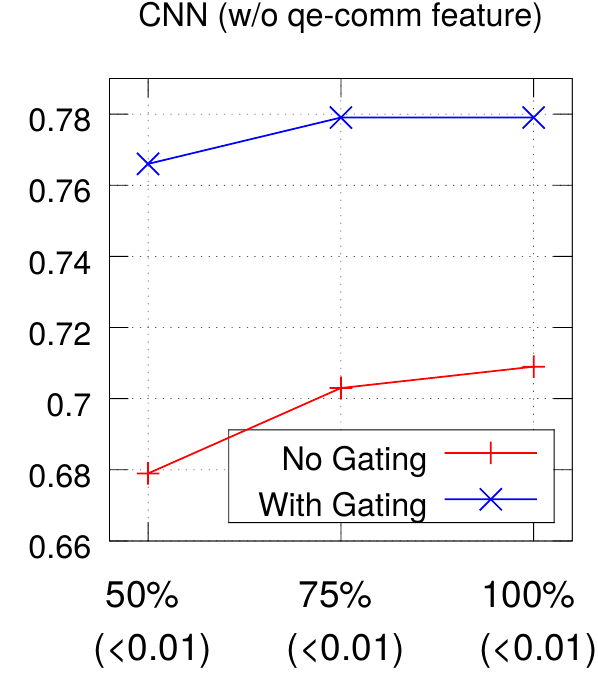}
        \label{fig:cnn_wo_f}
    \end{subfigure}
    \begin{subfigure}[b]{0.245\textwidth}
        \includegraphics[width=1.0\linewidth]{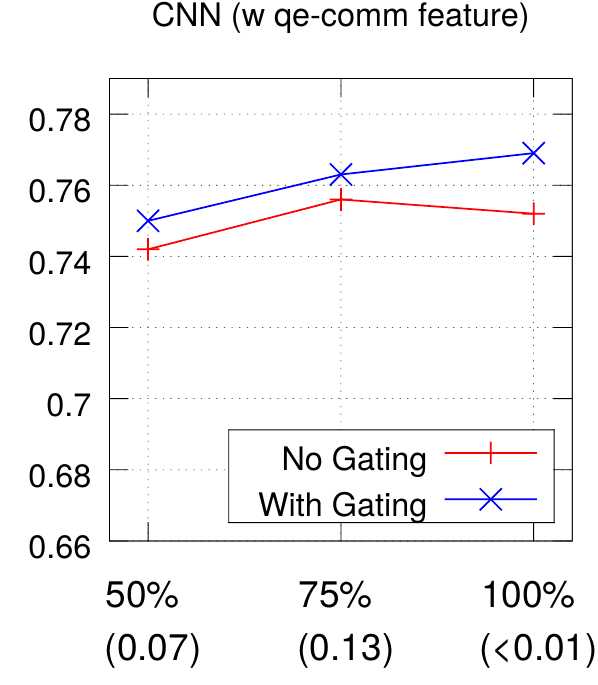}
        \label{fig:cnn_w_f}
    \end{subfigure}
        \begin{subfigure}[b]{0.245\textwidth}
    \includegraphics[width=1.0\linewidth]{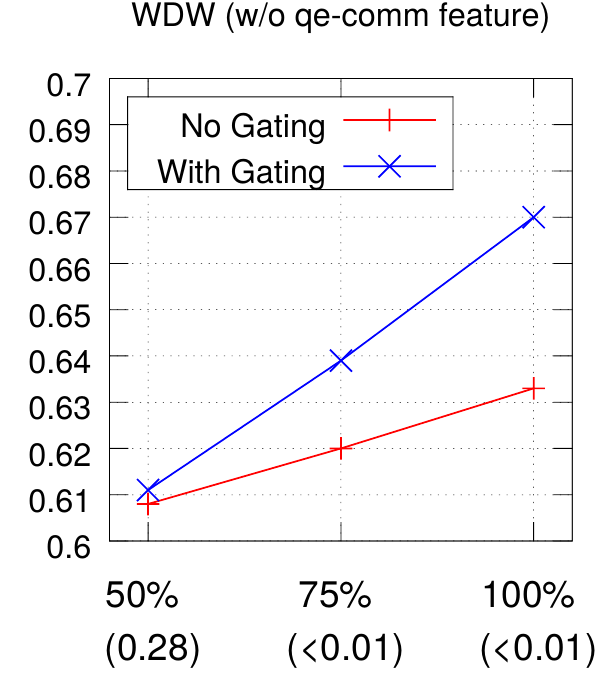}
        \label{fig:wdw_wo_f}
    \end{subfigure}
    \begin{subfigure}[b]{0.245\textwidth}
        \includegraphics[width=1.0\linewidth]{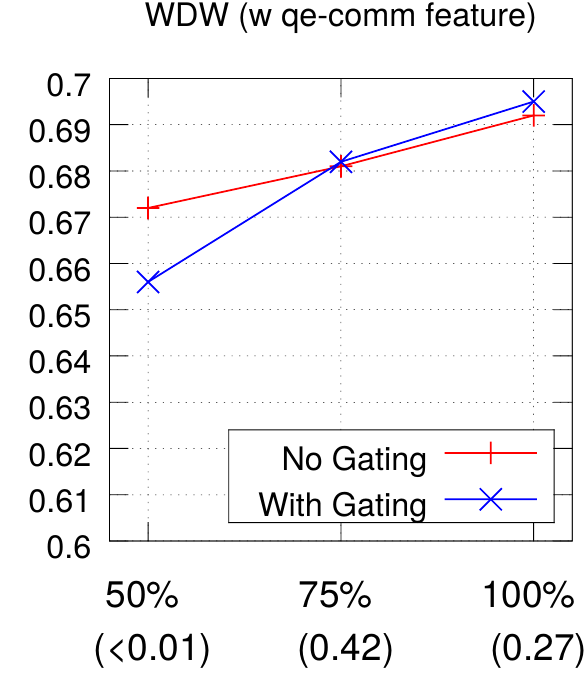}
        \label{fig:wdw_w_f}
    \end{subfigure}
    \label{fig:ablation}
\end{figure*}

In this section we do an ablation study to see the effect of Gated Attention. We compare the GA Reader as described here to a model which is exactly the same in all aspects, except that it passes document embeddings $D^{(k)}$ in each layer directly to the inputs of the next layer without using the GA module. In other words $X^{(k)}=D^{(k)}$ for all $k>0$. This model ends up using only one query GRU at the output layer for selecting the answer from the document. We compare these two variants both with and without the qe-comm feature on CNN and WDW datasets for three subsets of the training data - 50\%, 75\% and 100\%. Test set accuracies for these settings are shown in Figure~\ref{fig:ablation}. On CNN when tested without feature engineering, we observe that GA provides a significant boost in performance compared to without GA. When tested with the feature it still gives an improvement, but the improvement is significant only with 100\% training data. On WDW-Strict, which is a third of the size of CNN, without the feature we see an improvement when using GA versus without using GA, which becomes significant as the training set size increases. When tested with the feature on WDW, for a small data size without GA does better than with GA, but as the dataset size increases they become equivalent. We conclude that GA provides a boost in the absence of feature engineering, or as the training set size increases.

Next we look at the question of how to gate intermediate document reader states from the query, i.e. what operation to use in equation \ref{eq:gating}. Table ~\ref{tab:gating_fn} (top) shows the performance on WDW dataset for three common choices -- \texttt{sum} ($x=d+q$), \texttt{concatenate} ($x=d \Vert q$) and \texttt{multiply} ($x=d \odot q$). Empirically we find element-wise multiplication does significantly better than the other two, which justifies our motivation to ``filter'' out document features which are irrelevant to the query.

At the bottom of Table~\ref{tab:gating_fn} we show the effect of varying the number of hops $K$ of the GA Reader on the final performance. We note that for $K=1$, our model is equivalent to the AS Reader without any GA modules. We see a steep and steady rise in accuracy as the number of hops is increased from $K=1$ to $3$, which remains constant beyond that. This is a common trend in machine learning as model complexity is increased, however we note that a multi-hop architecture is important to achieve a high performance for this task, and provide further evidence for this in the next section.

\subsection{Ablation Study for Model Components}
\label{app:ablation}
Table~\ref{tab:ablation} shows accuracy on WDW by removing one component at a time.
\begin{table}[!htbp]
\centering
\caption{\small Ablation study on WDW dataset, without using the qe-comm feature and with fixed $L(w)$. Results marked with $\dagger$ are cf \citet{onishi2016did}.}
\label{tab:ablation}
\begin{tabular}{@{}l|c|c@{}}
\toprule
\multirow{2}{*}{\textbf{Model}} & \multicolumn{2}{c}{\textbf{Accuracy}} \\ \cmidrule(l){2-3} 
                                          & Val                & Test              \\ \midrule
GA                                       & \textbf{68.3}               & \textbf{68.0}             \\
\quad $-$char                               & 66.9               & 66.9              \\
\quad $-$token-attentions (eq. \ref{eq:q_att})  & 65.7               & 65.0              \\
\quad $-$glove, $+$corpus                & 64.0               & 62.5              \\
 \midrule
GA-{}-$\dagger$                & --               & 57              \\ \bottomrule
\end{tabular}
\end{table}
The steepest reduction is observed when we replace pretrained GloVe vectors with those pretrained on the corpus itself. GloVe vectors were trained on a large corpus of about 6 billion tokens \citep{pennington2014glove}, and provide an important source of prior knowledge for the model. Note that the strongest baseline on WDW, NSE~\citep{munkhdalai2016reasoning}, also uses pretrained GloVe vectors, hence the comparison is fair in that respect. Next, we observe a substantial drop when removing token-specific attentions over the query in the GA module, which allow gating individual tokens in the document only by parts of the query relevant to that token rather than the overall query representation. Finally, removing the character embeddings, which were only used for WDW and CBT, leads to a reduction of about 1\% in the performance. 


\subsection{Attention Visualization}
\begin{figure*}[t]
\centering
\caption{Layer-wise attention visualization of GA Reader trained on WDW-Strict. See text for details.}
\includegraphics[width=0.91\linewidth]{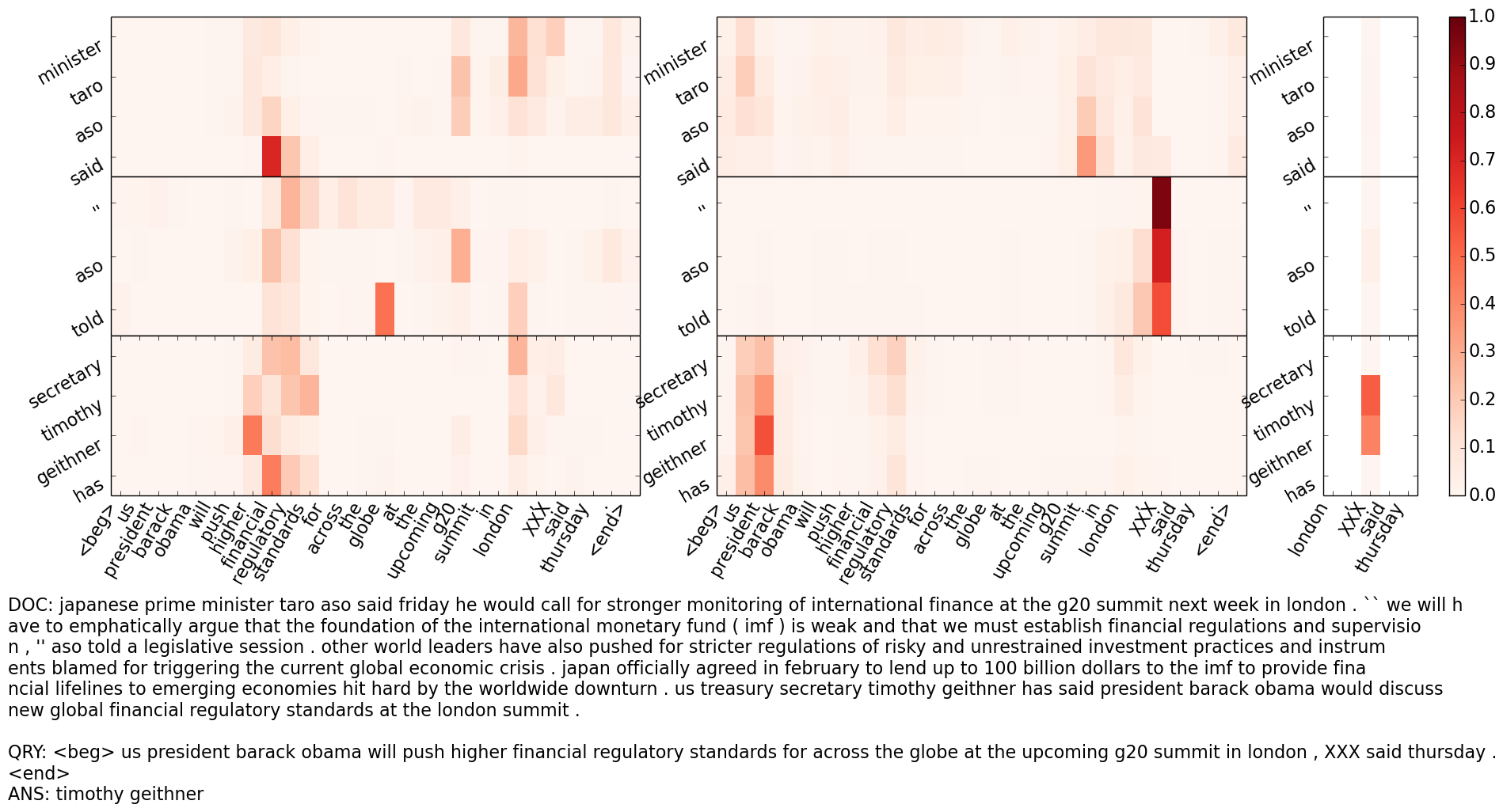}
\label{fig:att_plots}
\end{figure*}

To gain an insight into the reading process employed by the model we analyzed the attention distributions at intermediate layers of the reader.
Figure~\ref{fig:att_plots} shows an example from the validation set of WDW dataset (several more are in the Appendix).
In each figure, the left and middle plots visualize attention over the query (equation~\ref{eq:q_att}) for candidates in the document after layers 1 \& 2 respectively. The right plot shows attention over candidates in the document of cloze placeholder (\texttt{XXX}) in the query at the final layer.
The full document, query and correct answer are shown at the bottom.

A generic pattern observed in these examples is that in intermediate layers,
candidates in the document (shown along rows) tend to pick out salient tokens in the query which provide clues about the cloze, and in the final layer the candidate with the highest match with these tokens is selected as the answer.
In Figure~\ref{fig:att_plots} there is a high attention of the correct answer on \texttt{financial} \texttt{regulatory} \texttt{standards} in the first layer, and on \texttt{us} \texttt{president} in the second layer. The incorrect answer, in contrast, only attends to one of these aspects, and hence receives a lower score in the final layer despite the n-gram overlap it has with the cloze token in the query.
Importantly, different layers tend to focus on different tokens in the query, supporting the hypothesis that the multi-hop architecture of GA Reader is able to combine distinct pieces of information to answer the query.

\section{Conclusion}
\label{sec:conclusion}
We presented the Gated-Attention reader for answering cloze-style questions over documents.
The GA reader features a novel multiplicative gating mechanism, combined with a multi-hop architecture.
Our model achieves the state-of-the-art performance on several large-scale benchmark datasets with more than 4\% improvements over competitive baselines. Our model design is backed up by an ablation study showing statistically significant improvements of using Gated Attention as information filters. We also showed empirically that multiplicative gating is superior to addition and concatenation operations for implementing gated-attentions, though a theoretical justification remains part of future research goals. Analysis of document and query attentions in intermediate layers of the reader further reveals that the model iteratively attends to different aspects of the query to arrive at the final answer. In this paper we have focused on text comprehension, but we believe that the Gated-Attention mechanism may benefit other tasks as well where multiple sources of information interact.


\section*{Acknowledgments}
This work was funded by NSF under CCF1414030 and Google Research.

\bibliography{iclr2017_conference}

\begin{thebibliography}{32}
\providecommand{\natexlab}[1]{#1}
\providecommand{\url}[1]{\texttt{#1}}
\expandafter\ifx\csname urlstyle\endcsname\relax
  \providecommand{\doi}[1]{doi: #1}\else
  \providecommand{\doi}{doi: \begingroup \urlstyle{rm}\Url}\fi

\bibitem[Bahdanau et~al.(2014)Bahdanau, Cho, and Bengio]{bahdanau2014neural}
Dzmitry Bahdanau, Kyunghyun Cho, and Yoshua Bengio.
\newblock Neural machine translation by jointly learning to align and
  translate.
\newblock \emph{arXiv preprint arXiv:1409.0473}, 2014.

\bibitem[Bajgar et~al.(2016)Bajgar, Kadlec, and
  Kleindienst]{bajgar2016embracing}
Ondrej Bajgar, Rudolf Kadlec, and Jan Kleindienst.
\newblock Embracing data abundance: Booktest dataset for reading comprehension.
\newblock \emph{arXiv preprint arXiv:1610.00956}, 2016.

\bibitem[Chen et~al.(2016)Chen, Bolton, and Manning]{chen2016thorough}
Danqi Chen, Jason Bolton, and Christopher~D Manning.
\newblock A thorough examination of the cnn/daily mail reading comprehension
  task.
\newblock \emph{ACL}, 2016.

\bibitem[Cho et~al.(2015)Cho, Van~Merri{\"e}nboer, Gulcehre, Bahdanau,
  Bougares, Schwenk, and Bengio]{cho2014learning}
Kyunghyun Cho, Bart Van~Merri{\"e}nboer, Caglar Gulcehre, Dzmitry Bahdanau,
  Fethi Bougares, Holger Schwenk, and Yoshua Bengio.
\newblock Learning phrase representations using rnn encoder-decoder for
  statistical machine translation.
\newblock \emph{ACL}, 2015.

\bibitem[Cui et~al.(2017)Cui, Chen, Wei, Wang, Liu, and Hu]{cui2016attention}
Yiming Cui, Zhipeng Chen, Si~Wei, Shijin Wang, Ting Liu, and Guoping Hu.
\newblock Attention-over-attention neural networks for reading comprehension.
\newblock \emph{ACL}, 2017.

\bibitem[Dhingra et~al.(2016)Dhingra, Zhou, Fitzpatrick, Muehl, and
  Cohen]{dhingra2016tweet2vec}
Bhuwan Dhingra, Zhong Zhou, Dylan Fitzpatrick, Michael Muehl, and William~W
  Cohen.
\newblock Tweet2vec: Character-based distributed representations for social
  media.
\newblock \emph{ACL}, 2016.

\bibitem[Hermann et~al.(2015)Hermann, Kocisky, Grefenstette, Espeholt, Kay,
  Suleyman, and Blunsom]{hermann2015teaching}
Karl~Moritz Hermann, Tomas Kocisky, Edward Grefenstette, Lasse Espeholt, Will
  Kay, Mustafa Suleyman, and Phil Blunsom.
\newblock Teaching machines to read and comprehend.
\newblock In \emph{Advances in Neural Information Processing Systems}, pp.\
  1684--1692, 2015.

\bibitem[Hill et~al.(2016)Hill, Bordes, Chopra, and Weston]{hill2015goldilocks}
Felix Hill, Antoine Bordes, Sumit Chopra, and Jason Weston.
\newblock The goldilocks principle: Reading children's books with explicit
  memory representations.
\newblock \emph{ICLR}, 2016.

\bibitem[Hochreiter \& Schmidhuber(1997)Hochreiter and
  Schmidhuber]{hochreiter1997long}
Sepp Hochreiter and J{\"u}rgen Schmidhuber.
\newblock Long short-term memory.
\newblock \emph{Neural computation}, 9\penalty0 (8):\penalty0 1735--1780, 1997.

\bibitem[Kadlec et~al.(2016)Kadlec, Schmid, Bajgar, and
  Kleindienst]{kadlec2016text}
Rudolf Kadlec, Martin Schmid, Ondrej Bajgar, and Jan Kleindienst.
\newblock Text understanding with the attention sum reader network.
\newblock \emph{ACL}, 2016.

\bibitem[Kingma \& Ba(2015)Kingma and Ba]{kingma2014adam}
Diederik Kingma and Jimmy Ba.
\newblock Adam: A method for stochastic optimization.
\newblock \emph{ICLR}, 2015.

\bibitem[Kiros et~al.(2014)Kiros, Zemel, and
  Salakhutdinov]{kiros2014multiplicative}
Ryan Kiros, Richard Zemel, and Ruslan~R Salakhutdinov.
\newblock A multiplicative model for learning distributed text-based attribute
  representations.
\newblock In \emph{Advances in Neural Information Processing Systems}, pp.\
  2348--2356, 2014.

\bibitem[Kobayashi et~al.(2016)Kobayashi, Tian, Okazaki, and
  Inui]{kobayashi2016dynamic}
Sosuke Kobayashi, Ran Tian, Naoaki Okazaki, and Kentaro Inui.
\newblock Dynamic entity representations with max-pooling improves machine
  reading.
\newblock In \emph{NAACL-HLT}, 2016.

\bibitem[Kumar et~al.(2016)Kumar, Irsoy, Su, Bradbury, English, Pierce,
  Ondruska, Gulrajani, and Socher]{kumar2015ask}
Ankit Kumar, Ozan Irsoy, Jonathan Su, James Bradbury, Robert English, Brian
  Pierce, Peter Ondruska, Ishaan Gulrajani, and Richard Socher.
\newblock Ask me anything: Dynamic memory networks for natural language
  processing.
\newblock \emph{ICML}, 2016.

\bibitem[Li et~al.(2016)Li, Li, He, Wang, Cao, Zhou, and Xu]{li2016dataset}
Peng Li, Wei Li, Zhengyan He, Xuguang Wang, Ying Cao, Jie Zhou, and Wei Xu.
\newblock Dataset and neural recurrent sequence labeling model for open-domain
  factoid question answering.
\newblock \emph{arXiv preprint arXiv:1607.06275}, 2016.

\bibitem[Mitchell \& Lapata(2008)Mitchell and Lapata]{mitchell2008vector}
Jeff Mitchell and Mirella Lapata.
\newblock Vector-based models of semantic composition.
\newblock In \emph{ACL}, pp.\  236--244, 2008.

\bibitem[Mnih et~al.(2014)Mnih, Heess, Graves, et~al.]{mnih2014recurrent}
Volodymyr Mnih, Nicolas Heess, Alex Graves, et~al.
\newblock Recurrent models of visual attention.
\newblock In \emph{Advances in Neural Information Processing Systems}, pp.\
  2204--2212, 2014.

\bibitem[Munkhdalai \& Yu(2017{\natexlab{a}})Munkhdalai and
  Yu]{munkhdalai2016neural}
Tsendsuren Munkhdalai and Hong Yu.
\newblock Neural semantic encoders.
\newblock \emph{EACL}, 2017{\natexlab{a}}.

\bibitem[Munkhdalai \& Yu(2017{\natexlab{b}})Munkhdalai and
  Yu]{munkhdalai2016reasoning}
Tsendsuren Munkhdalai and Hong Yu.
\newblock Reasoning with memory augmented neural networks for language
  comprehension.
\newblock \emph{ICLR}, 2017{\natexlab{b}}.

\bibitem[Onishi et~al.(2016)Onishi, Wang, Bansal, Gimpel, and
  McAllester]{onishi2016did}
Takeshi Onishi, Hai Wang, Mohit Bansal, Kevin Gimpel, and David McAllester.
\newblock Who did what: A large-scale person-centered cloze dataset.
\newblock \emph{EMNLP}, 2016.

\bibitem[Pascanu et~al.(2013)Pascanu, Mikolov, and
  Bengio]{pascanu2012difficulty}
Razvan Pascanu, Tomas Mikolov, and Yoshua Bengio.
\newblock On the difficulty of training recurrent neural networks.
\newblock \emph{ICML (3)}, 28:\penalty0 1310--1318, 2013.

\bibitem[Pennington et~al.(2014)Pennington, Socher, and
  Manning]{pennington2014glove}
Jeffrey Pennington, Richard Socher, and Christopher~D. Manning.
\newblock Glove: Global vectors for word representation.
\newblock In \emph{Empirical Methods in Natural Language Processing (EMNLP)},
  pp.\  1532--1543, 2014.
\newblock URL \url{http://www.aclweb.org/anthology/D14-1162}.

\bibitem[Seo et~al.(2017)Seo, Kembhavi, Farhadi, and
  Hajishirzi]{seo2016bidirectional}
Minjoon Seo, Aniruddha Kembhavi, Ali Farhadi, and Hannaneh Hajishirzi.
\newblock Bidirectional attention flow for machine comprehension.
\newblock \emph{ICLR}, 2017.

\bibitem[Shen et~al.(2016)Shen, Huang, Gao, and Chen]{shen2016reasonet}
Yelong Shen, Po-Sen Huang, Jianfeng Gao, and Weizhu Chen.
\newblock Reasonet: Learning to stop reading in machine comprehension.
\newblock \emph{arXiv preprint arXiv:1609.05284}, 2016.

\bibitem[Sordoni et~al.(2016)Sordoni, Bachman, and
  Bengio]{sordoni2016iterative}
Alessandro Sordoni, Phillip Bachman, and Yoshua Bengio.
\newblock Iterative alternating neural attention for machine reading.
\newblock \emph{arXiv preprint arXiv:1606.02245}, 2016.

\bibitem[Sukhbaatar et~al.(2015)Sukhbaatar, Weston, Fergus,
  et~al.]{sukhbaatar2015end}
Sainbayar Sukhbaatar, Jason Weston, Rob Fergus, et~al.
\newblock End-to-end memory networks.
\newblock In \emph{Advances in Neural Information Processing Systems}, pp.\
  2431--2439, 2015.

\bibitem[{Theano Development Team}(2016)]{2016arXiv160502688short}
{Theano Development Team}.
\newblock {Theano: A {Python} framework for fast computation of mathematical
  expressions}.
\newblock \emph{arXiv e-prints}, abs/1605.02688, May 2016.
\newblock URL \url{http://arxiv.org/abs/1605.02688}.

\bibitem[Trischler et~al.(2016)Trischler, Ye, Yuan, and
  Suleman]{trischler2016natural}
Adam Trischler, Zheng Ye, Xingdi Yuan, and Kaheer Suleman.
\newblock Natural language comprehension with the epireader.
\newblock \emph{EMNLP}, 2016.

\bibitem[Weston et~al.(2015)Weston, Chopra, and Bordes]{weston2014memory}
Jason Weston, Sumit Chopra, and Antoine Bordes.
\newblock Memory networks.
\newblock \emph{ICLR}, 2015.

\bibitem[Wu et~al.(2016)Wu, Zhang, Zhang, Bengio, and
  Salakhutdinov]{wu2016multiplicative}
Yuhuai Wu, Saizheng Zhang, Ying Zhang, Yoshua Bengio, and Ruslan Salakhutdinov.
\newblock On multiplicative integration with recurrent neural networks.
\newblock \emph{Advances in Neural Information Processing Systems}, 2016.

\bibitem[Yang et~al.(2014)Yang, Yih, He, Gao, and Deng]{yang2014learning}
Bishan Yang, Wen-tau Yih, Xiaodong He, Jianfeng Gao, and Li~Deng.
\newblock Learning multi-relational semantics using neural-embedding models.
\newblock \emph{NIPS Workshop on Learning Semantics}, 2014.

\bibitem[Yang et~al.(2016)Yang, Salakhutdinov, and Cohen]{yang2016multi}
Zhilin Yang, Ruslan Salakhutdinov, and William Cohen.
\newblock Multi-task cross-lingual sequence tagging from scratch.
\newblock \emph{arXiv preprint arXiv:1603.06270}, 2016.

\end{thebibliography}
\bibliographystyle{iclr2017_conference}

\appendix

\vspace{-0.05in}
\section{Implementation Details}
\label{app:implementation}
\begin{table*}[t]
\centering
\caption{\small Dataset statistics.}
\label{tab:data}
\begin{tabular}{@{}crrrrrr@{}}
\toprule
 & \textbf{CNN} & \textbf{Daily Mail} & \textbf{CBT-NE} & \textbf{CBT-CN} & \textbf{WDW-Strict} & \textbf{WDW-Relaxed} \\ \midrule
\# train              & 380,298             & 879,450               & 108,719              & 120,769  & 	 127,786	&	185,978           \\
\# validation        & 3,924             & 64,835              & 2,000             & 2,000  	&	10,000	&	10,000           \\
\# test           &  3,198            & 53,182               & 2,500              & 2,500  	&	10,000	&	10,000            \\
\# vocab           & 118,497             & 208,045               & 53,063              & 53,185 	&	347,406	&	 308,602             \\ 
max doc length	& 2,000	&	2,000 &	1,338 &	1,338 & 3,085	&	3,085 \\ \bottomrule
\end{tabular}
\end{table*}

\begin{table*}[t]
\centering
\caption{\small Hyperparameter settings for each dataset. dim() indicates hidden state size of GRU.}
\label{tab:params}
\begin{tabular}{@{}ccccccc@{}}
\toprule
\textbf{Hyperparameter} & \textbf{CNN} & \textbf{Daily Mail} & \textbf{CBT-NE} & \textbf{CBT-CN} & \textbf{WDW-Strict} & \textbf{WDW-Relaxed} \\ \midrule
Dropout	&	0.2	&	0.1	&	0.4	&	0.4	&	0.3	&	0.3	\\
$\mathrm{dim}(\bigru_*)$ &	256	&	256	&	128	&	128	&	128	&	128	\\ \bottomrule
\end{tabular}
\end{table*}

Our model was implemented using the Theano \citep{2016arXiv160502688short} and Lasagne\footnote{\scriptsize \url{https://lasagne.readthedocs.io/en/latest/}} Python libraries. We used stochastic gradient descent with ADAM updates for optimization, which combines classical momentum and adaptive gradients \citep{kingma2014adam}. The batch size was 32 and the initial learning rate was $5\times 10^{-4}$ which was halved every epoch after the second epoch.
The same setting is applied to all models and datasets. We also used gradient clipping with a threshold of 10 to stabilize GRU training \citep{pascanu2012difficulty}.
We set the number of layers $K$ to be $3$ for all experiments.
The number of hidden units for the character GRU was set to 50.
The remaining two hyperparameters---size of document and query GRUs, and dropout rate---were tuned on the validation set, and their optimal values are shown in Table~\ref{tab:params}. In general, the optimal GRU size increases and the dropout rate decreases as the corpus size increases.

The word lookup table was initialized with $100d$ GloVe vectors\footnote{\scriptsize \url{http://nlp.stanford.edu/projects/glove/}} \citep{pennington2014glove} and OOV tokens at test time were assigned unique random vectors. We empirically observed that initializing with pre-trained embeddings gives higher performance compared to random initialization for all datasets. Furthermore, for smaller datasets (WDW and CBT) we found that fixing these embeddings to their pretrained values led to higher test performance, possibly since it avoids overfitting. We do not use the character composition model for CNN and Daily Mail, since their entities (and hence candidate answers) are anonymized to generic tokens. For other datasets the character lookup table was randomly initialized with $25d$ vectors. All other parameters were initialized to their default values as specified in the Lasagne library.

\section{Attention Plots}
\begin{figure*}[h]
\centering
\caption{Layer-wise attention visualization of GA Reader trained on WDW-Strict. See text for details.}
\begin{subfigure}[b]{\textwidth}
\centering
\includegraphics[width=\linewidth]{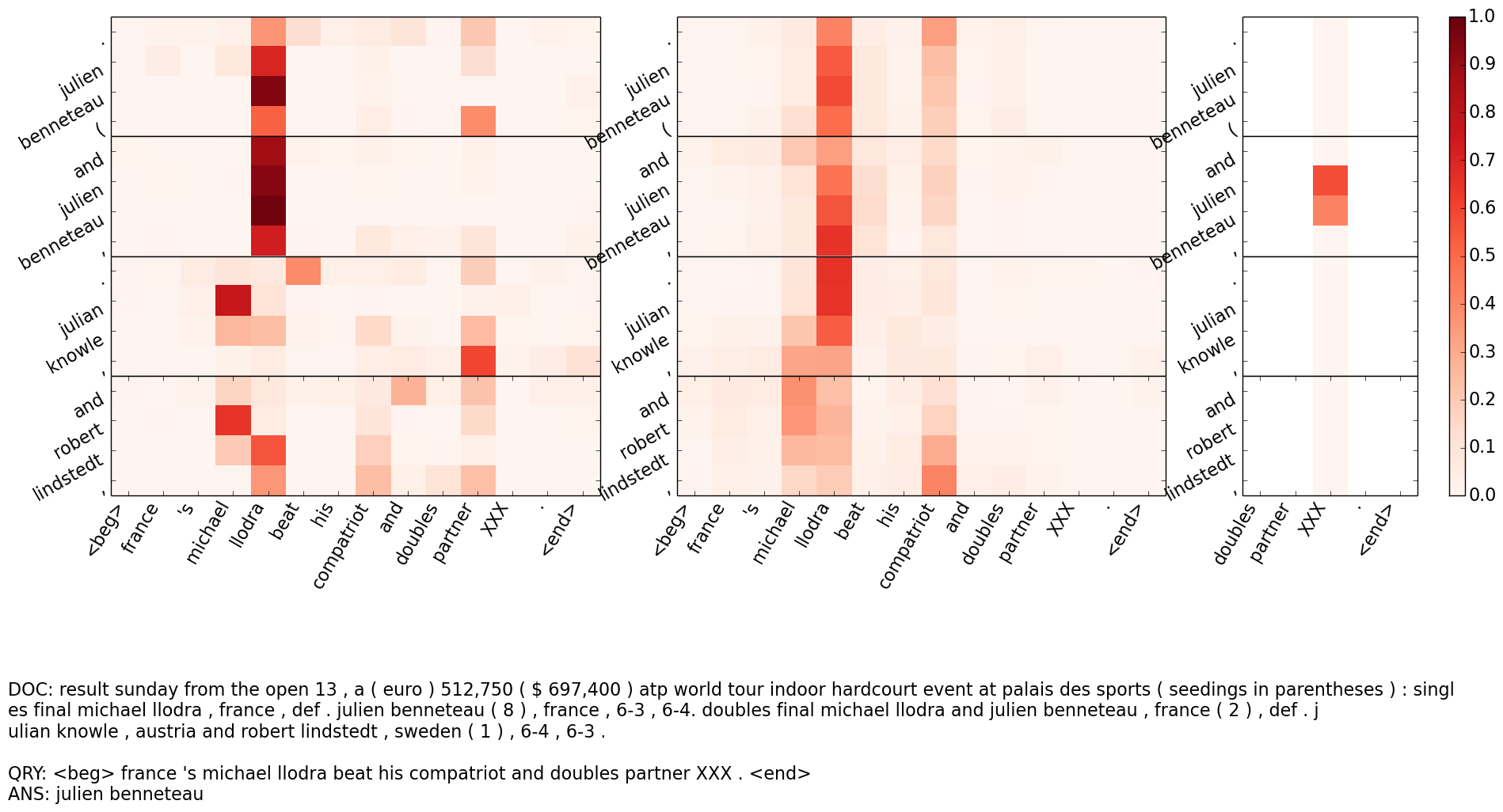}
\end{subfigure}
\begin{subfigure}[b]{\textwidth}
\centering
\includegraphics[width=\linewidth]{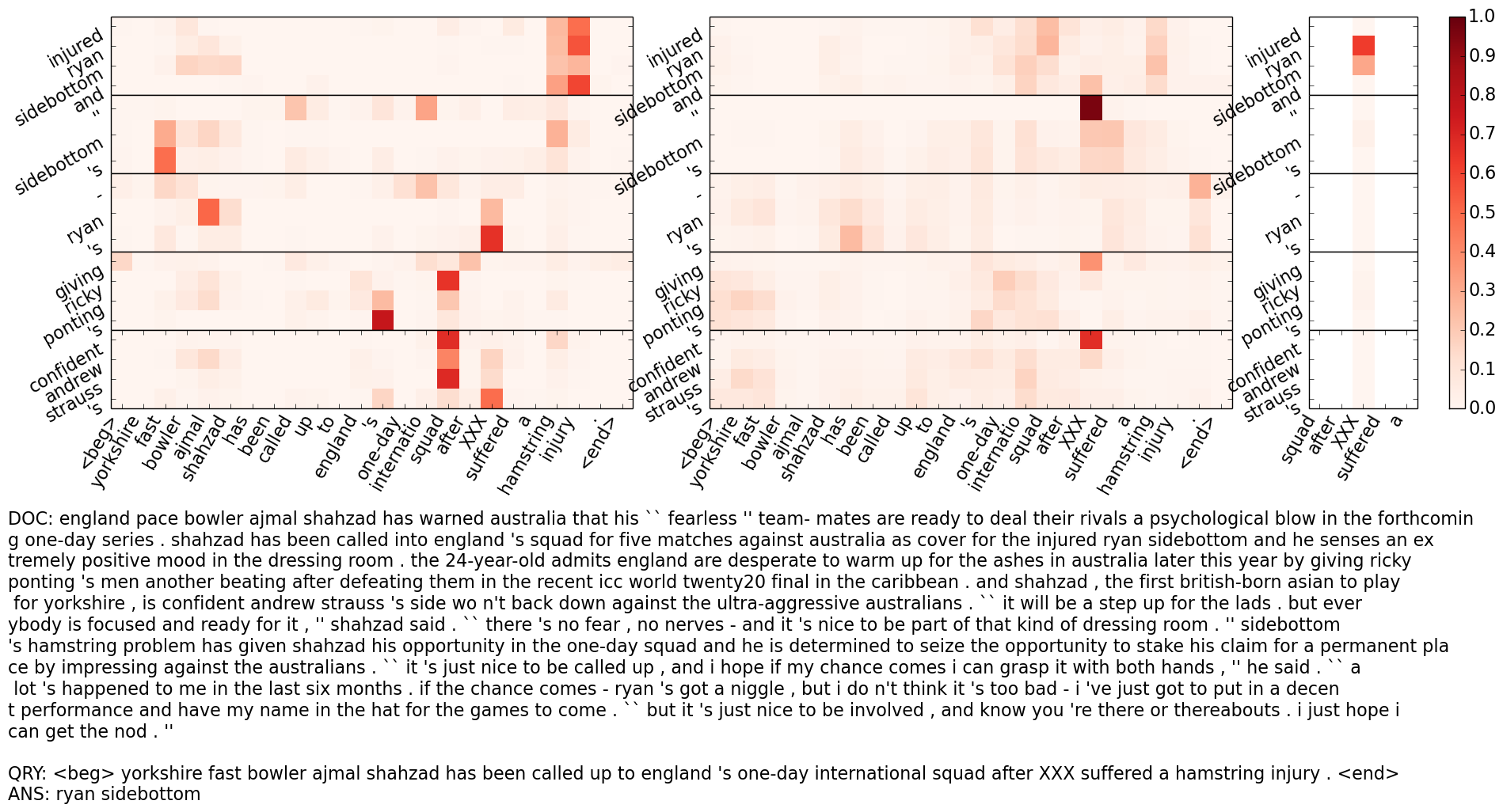}
\end{subfigure}
\end{figure*}

\begin{figure*}[h]
\centering
\caption{Layer-wise attention visualization of GA Reader trained on WDW-Strict. See text for details.}
\begin{subfigure}[b]{\textwidth}
\centering
\includegraphics[width=\linewidth]{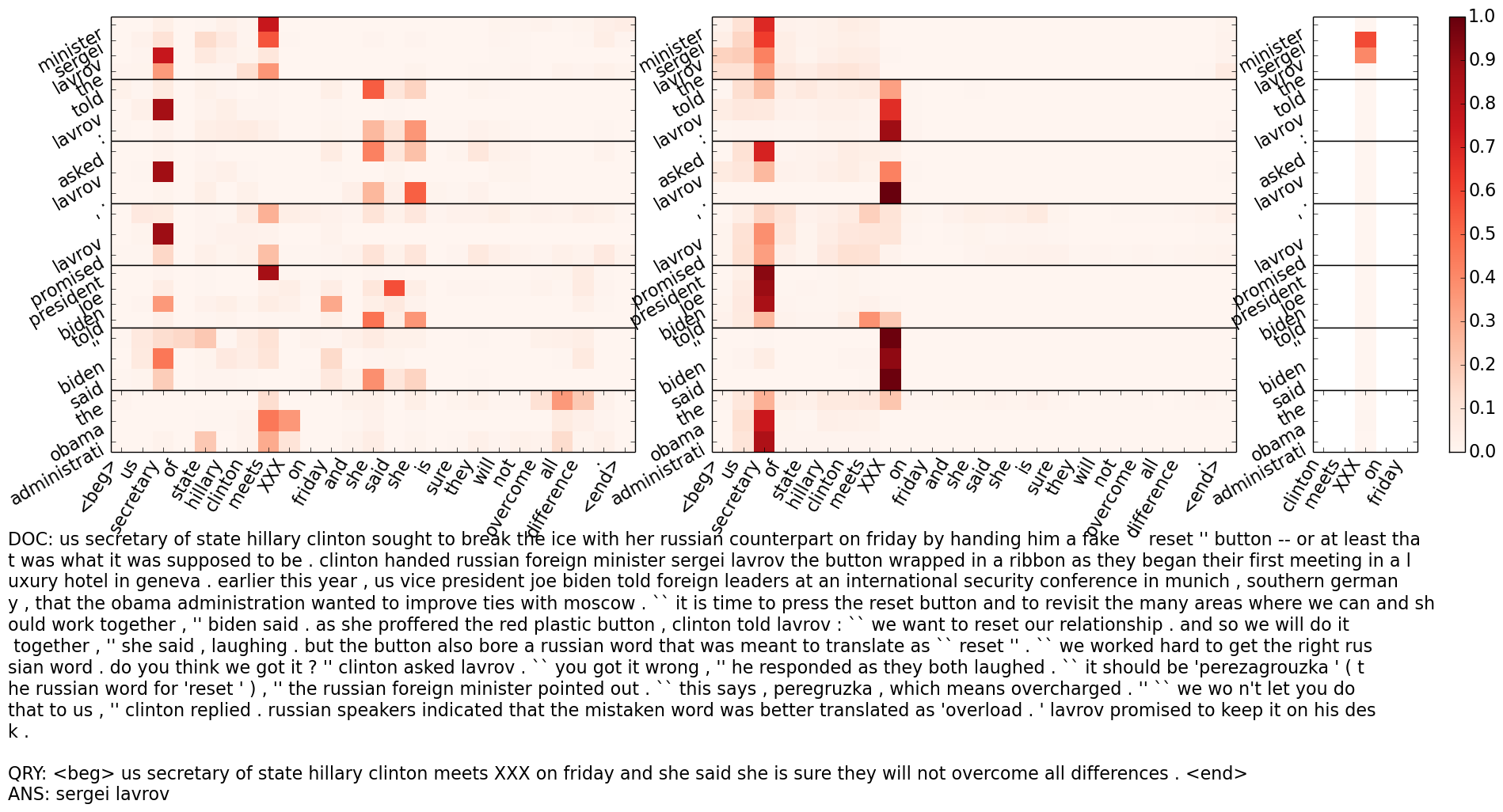}
\end{subfigure}
\begin{subfigure}[b]{\textwidth}
\includegraphics[width=\linewidth]{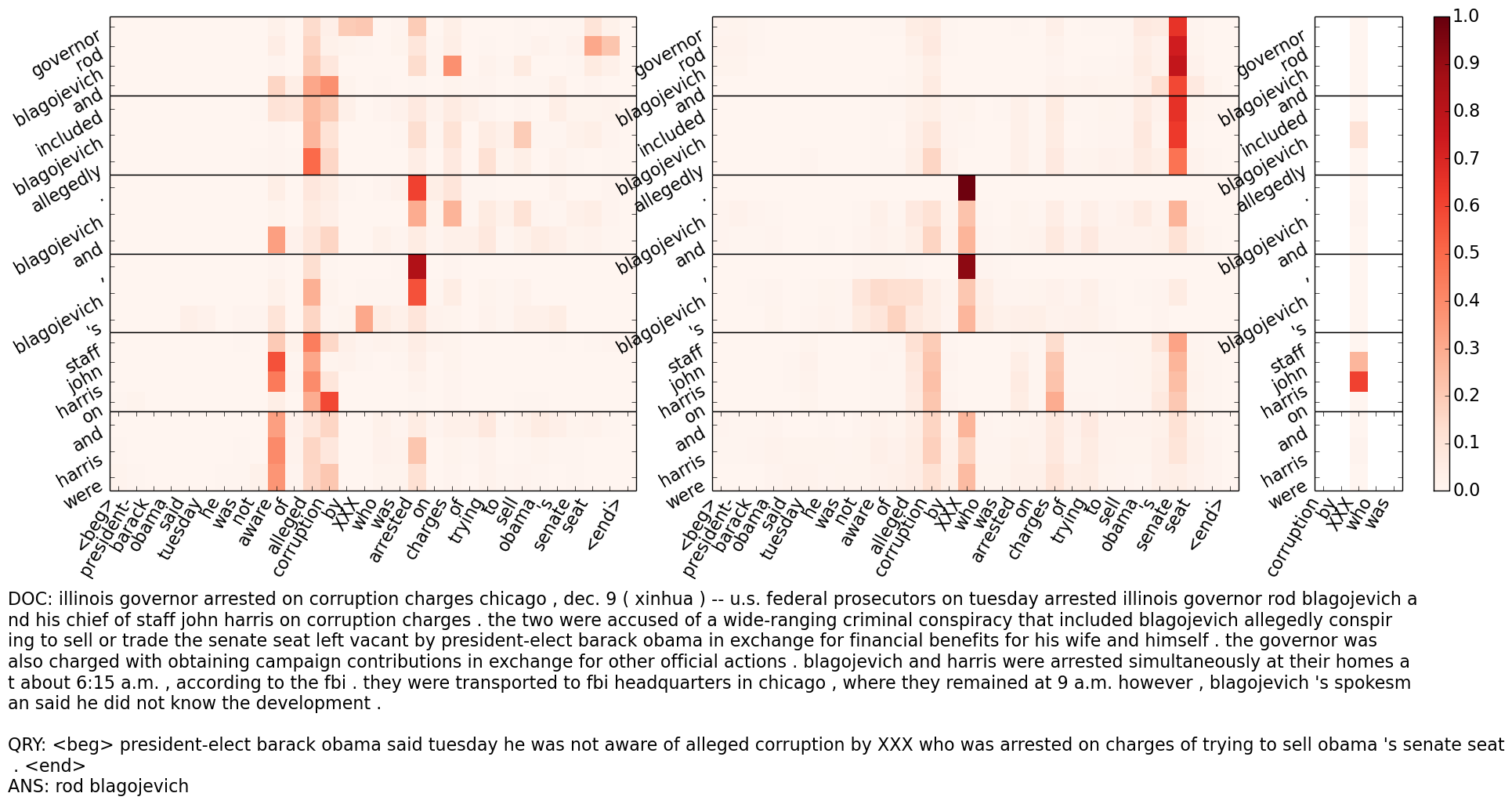}
\end{subfigure}
\end{figure*}

\begin{figure*}
\centering
\caption{\small Layer-wise attention visualization of GA Reader trained on WDW-Strict. See text for details.}
\begin{subfigure}[b]{\textwidth}
\centering
\includegraphics[width=\linewidth]{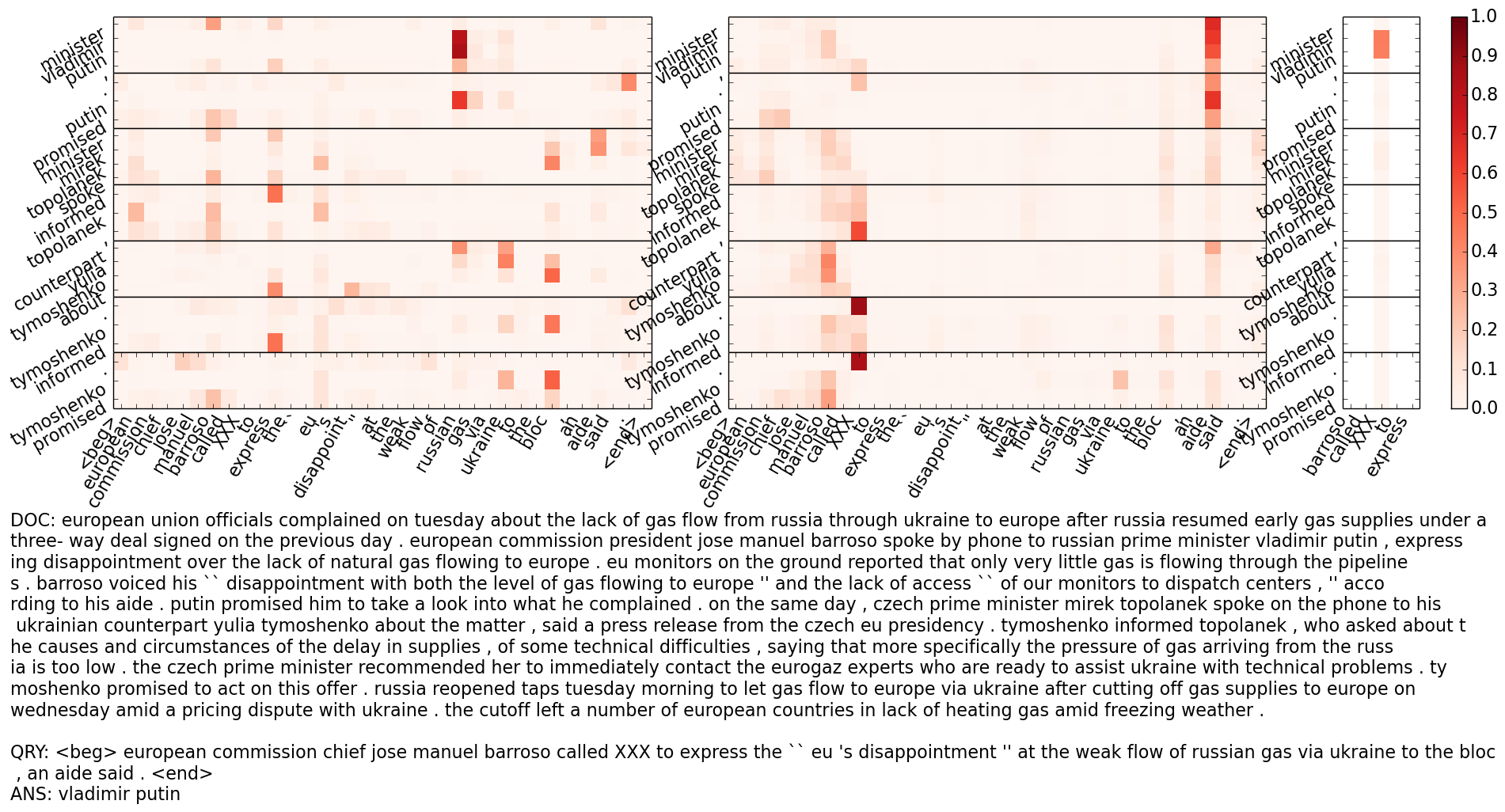}
\end{subfigure}
\begin{subfigure}[b]{\textwidth}
\centering
\includegraphics[width=\linewidth]{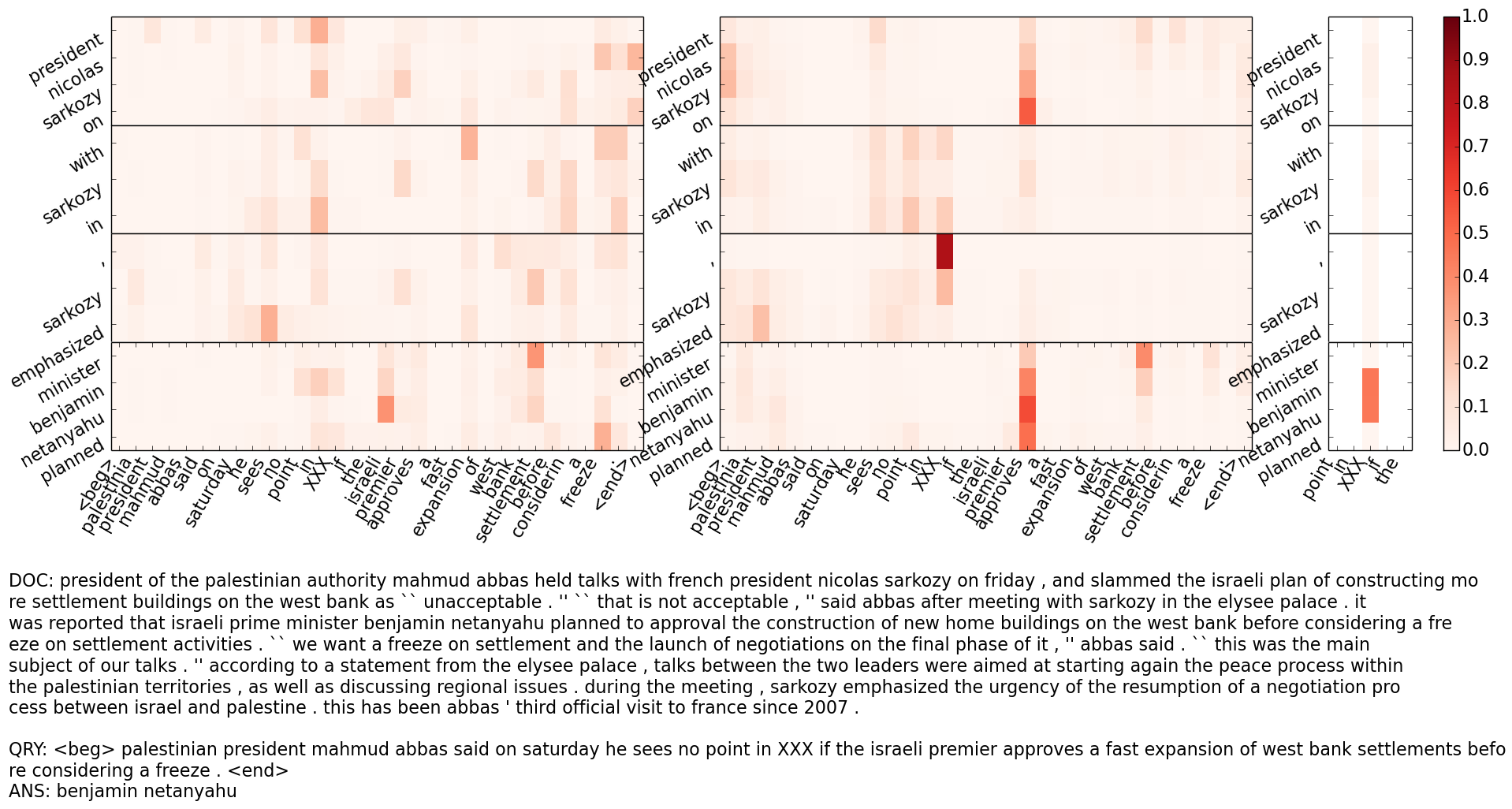}
\end{subfigure}
\end{figure*}

\begin{figure*}
\centering
\caption{\small Layer-wise attention visualization of GA Reader trained on WDW-Strict. See text for details.}
\begin{subfigure}[b]{\textwidth}
\centering
\includegraphics[width=\linewidth]{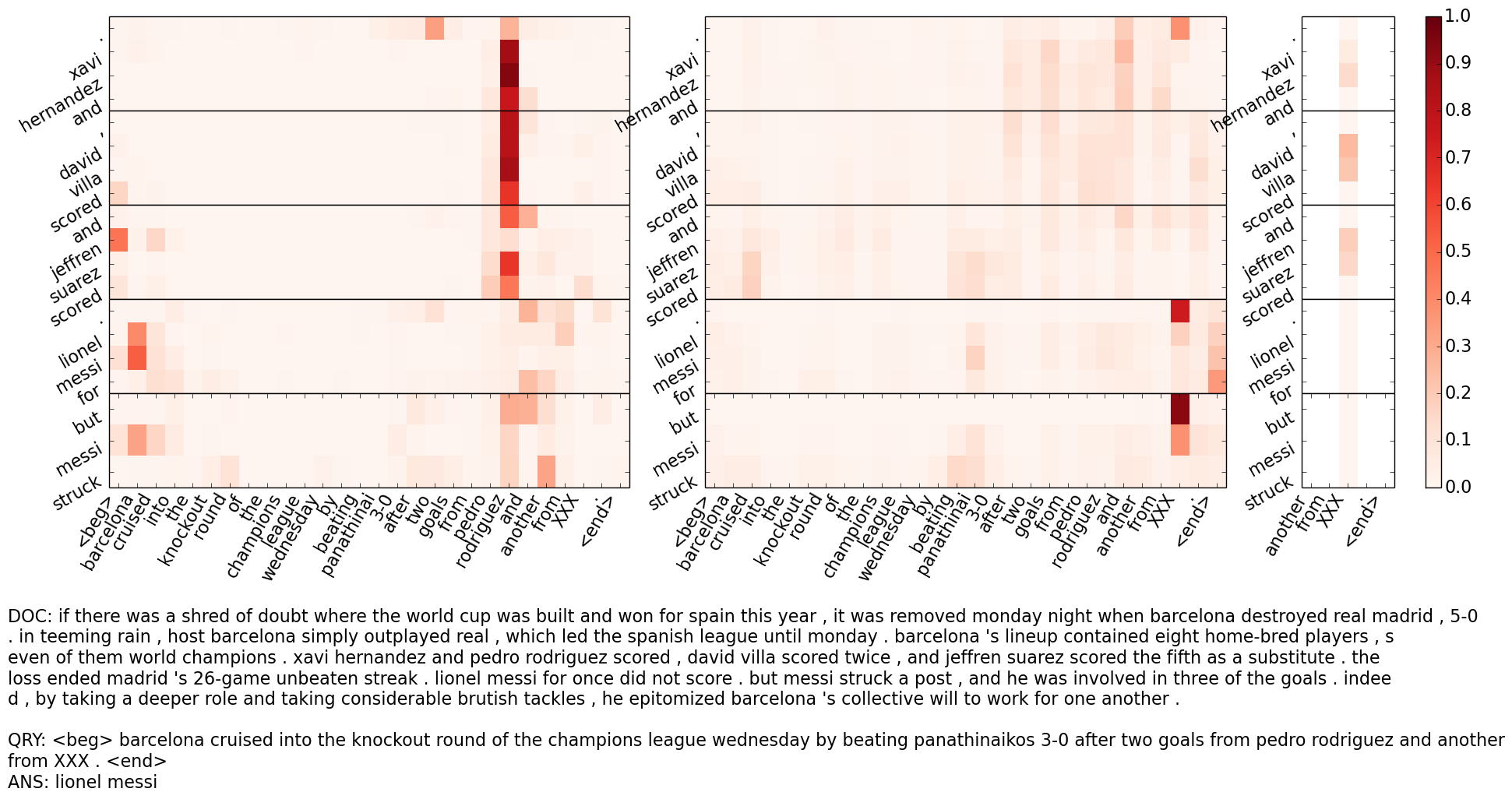}
\end{subfigure}
\begin{subfigure}[b]{\textwidth}
\centering
\includegraphics[width=\linewidth]{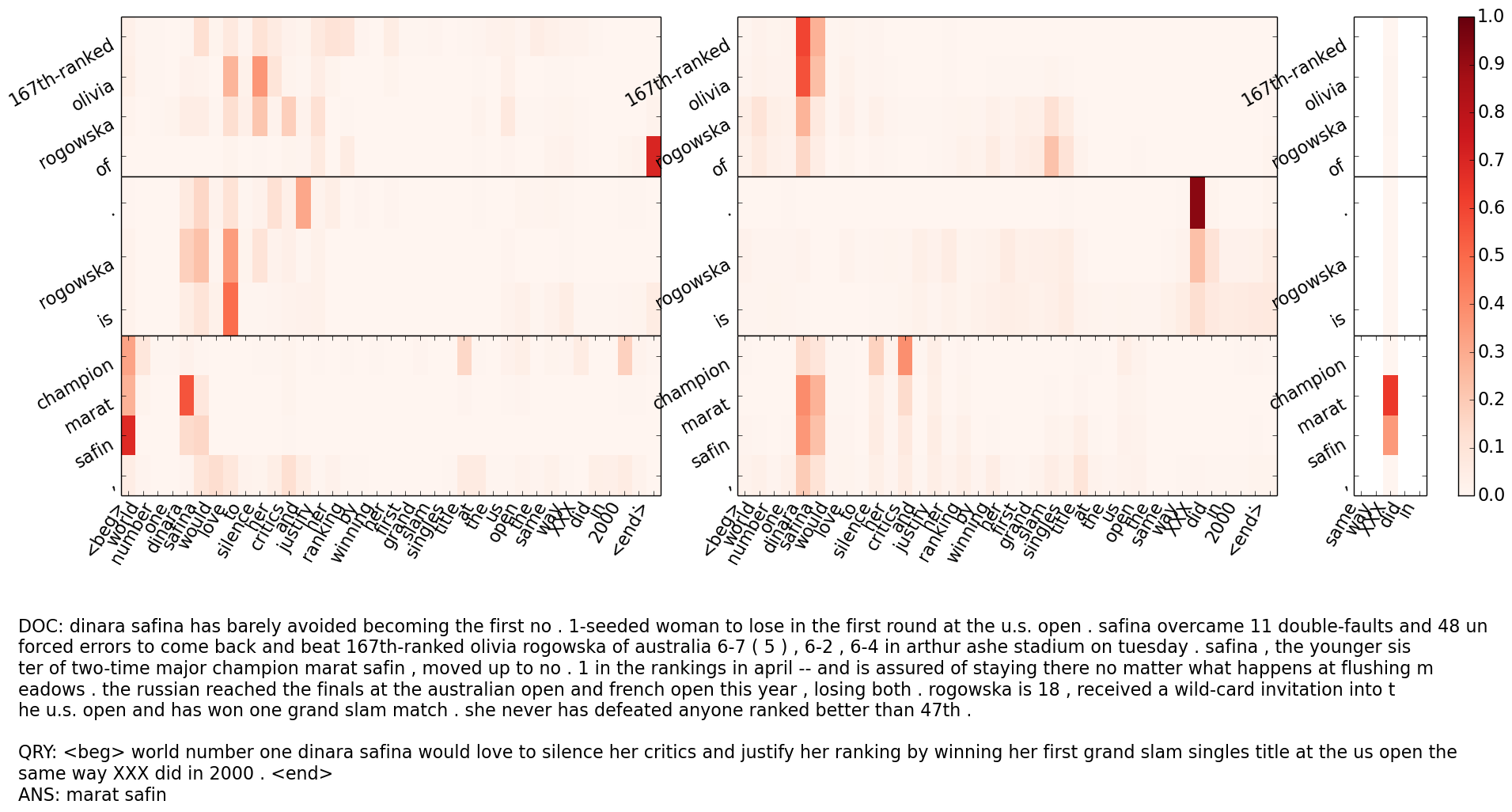}
\end{subfigure}
\end{figure*}

\end{document}